%% file: main.tex
\documentclass[conference]{IEEEtran}
\IEEEoverridecommandlockouts
\usepackage{cite}
\usepackage{amsmath,amssymb,amsfonts}
\usepackage{algorithmic}
\usepackage{graphicx}
\usepackage{textcomp}
\usepackage{multirow}
\usepackage{makecell}
\usepackage{xcolor}
\usepackage{booktabs}
\usepackage{array}
\usepackage[caption=false,font=normalsize,labelfont=sf,textfont=sf]{subfig}
\usepackage{textcomp}
\usepackage{stfloats}
\usepackage{url}
\usepackage{verbatim}
\usepackage{tikz}
\usepackage{float}
\usepackage{caption}
\usepackage[pagebackref=false,breaklinks=true,colorlinks=true,bookmarks=false,citecolor=blue,linkcolor=red]{hyperref}
\usepackage{cleveref}
\crefname{section}{Sect.}{Sects.}
\crefname{figure}{Fig.}{Figs.}
\crefname{table}{Tab.}{Tabs.}

\def\BibTeX{{\rm B\kern-.05em{\sc i\kern-.025em b}\kern-.08em
    T\kern-.1667em\lower.7ex\hbox{E}\kern-.125emX}}

\title{Spatiotemporally Decoupled Autoregressive Diffusion Model for Human Motion Generation} 
\author{Chengqun~Yang, Liang~Xu, Yanping~Li, Fulong~Liu, Jingnan~Gao, Weili~Zeng, Yichao~Yan\textsuperscript{*}
\\
MoE Key Lab of Artificial Intelligence, AI Institute, Shanghai Jiao Tong University
\\
{\tt\small \{ycq0191, liangxu, ypli2024, lfl925884898, gjn0310, zwl666, yanyichao\}@sjtu.edu.cn}
}

\begin{document}

\maketitle

\renewcommand{\thefootnote}{}
\renewcommand{\footnoterule}{%
  \vspace*{-3pt} 
  \hrule width 2in height 0.4pt 
  \vspace*{2pt} 
}

\footnotetext{%
\parbox[t]{2in}{%
\hbox to \linewidth{{*} Corresponding author: Yichao Yan.\hfill}%
\hbox to \linewidth{This work was supported in part by NSFC (62571314).\hfill}%
}}
\input{sections/1_intro_v1}

\input{sections/2_related_works_v1}

\input{sections/3_method_v1}
\input{sections/4_experiment_v1}
\input{sections/5_conclusion_v1}

\bibliographystyle{IEEEbib}
\bibliography{reference}

\end{document}

%% file: sections/1_intro_v1.tex
\begin{abstract}
Text-driven human motion synthesis has made substantial development with two core modules of motion representation and generative architecture. For representation, Vector Quantization (VQ)-based methods compress motion data into discrete tokens while latent-based models operate directly in continuous space.
However, both of these representations exhibit significant limitations. VQ-based methods suffer from inherent information loss, which compromises the quality, diversity, and generalization of generated motions, while continuous representation on holistic whole-body motion hinders part-level flexibility.
For architecture, diffusion and autoregressive diffusion models have demonstrated their superiority, yet the fine-grained controllability over individual body parts is also limited.
Thus, we propose a unified spatiotemporally decoupled framework named DeMoDiff, which jointly redesigns representation and architecture. 
To enhance representation extraction capabilities and offer greater part-level controllability, we present a spatial-temporal VAE that encodes each body joint rather than compressing the whole-body motion into a single latent space.
Then, we incorporate spatial-temporal masking and attention mechanisms into an autoregressive diffusion generator, achieving both generative capability and controllable editability.
Extensive experiments on the HumanML3D and KIT-ML datasets demonstrate that our model achieves state-of-the-art reconstruction performance and compelling motion generation results.
Moreover, our framework demonstrates strong temporal and spatial editing capabilities, further validating its effectiveness. 
Our project page: \href{https://rex0191.github.io/DeMoDiff/}{https://rex0191.github.io/DeMoDiff/}.

\end{abstract}

\begin{IEEEkeywords}
Text-driven human motion generation, Autoregressive diffusion model, Spatial-temporal decoupling.
\end{IEEEkeywords}

\section{Introduction}

Text-driven human motion generation~\cite{MDM,MLD,mogents,momask, actiongpt} is a rapidly advancing frontier in computer vision, focusing on synthesizing realistic and controllable 3D human motion sequences that semantically align with the given language prompt. 
Existing approaches for human motion generation typically comprise two essential components: motion representation and generative architecture.

Regarding motion representation, mainstream approaches primarily fall into two dominant paradigms of Vector Quantization (VQ)-based~\cite{TM2T,T2MGPT,motiongpt,momask,mogents} and latent-based representations~\cite{mardm,MLD,motionstreamer}.
Generative architectures are inherently entangled with motion representations. For instance, VQ-based methods typically employ autoregressive models to predict discrete tokens~\cite{T2MGPT,momask,mogents}.
Conversely, diffusion~\cite{MLD, MDM, MotionDiffuse, remodiffuse} and autoregressive diffusion models~\cite{motionstreamer, mardm} have demonstrated their superiority, yet the fine-grained controllability over individual body parts is also limited.
Although continuous regression avoids the quantization-induced precision loss of VQ-based methods, it introduces a more challenging problem of directly regressing articulated and high-dimensional human motions, making fine-grained manipulation infeasible.
\begin{figure} 
    \includegraphics[width=\columnwidth]{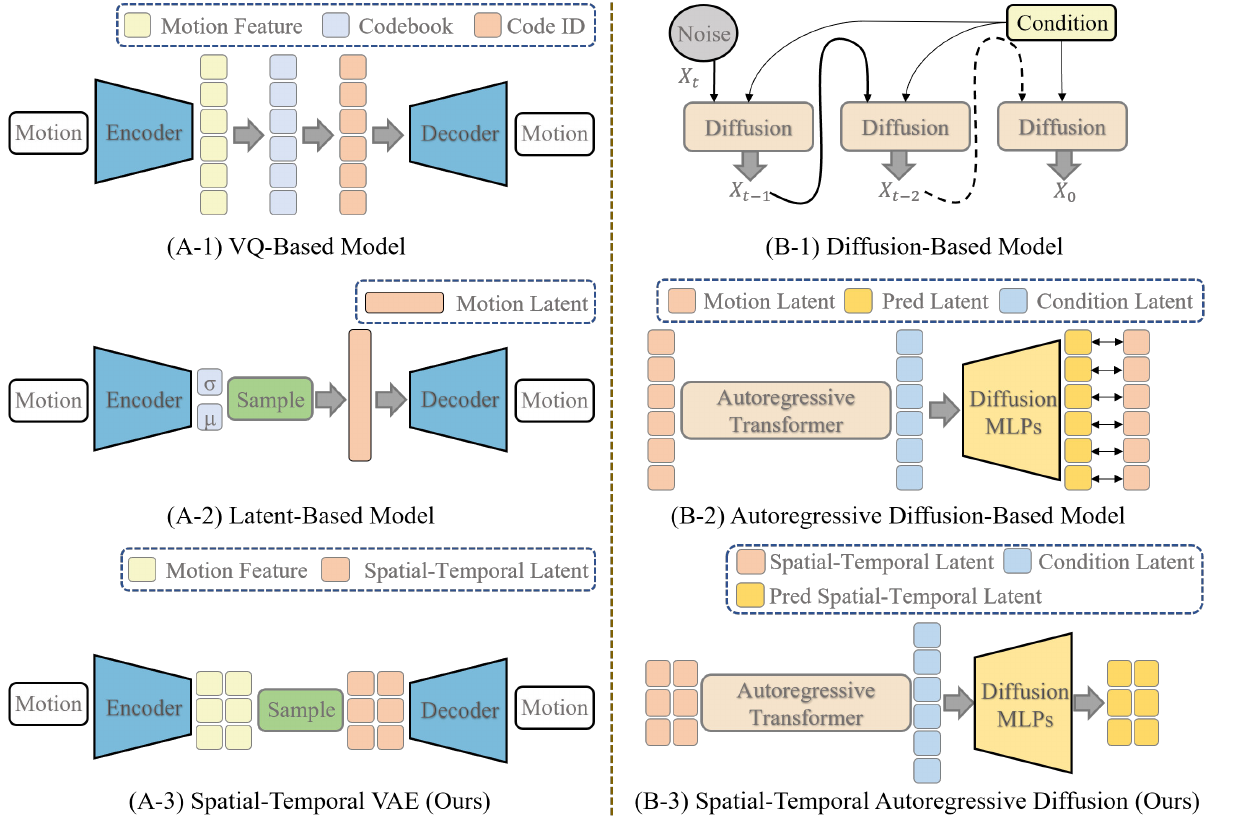}
    \caption{\textbf{Comparison of motion representation (A1-3) and generation model (B1-3).} Our proposed DeMoDiff is composed of spatial-temporal VAE and spatial-temporal autoregressive diffusion model for enhanced motion representation, generation, and spatiotemporal editing capabilities.}
  \label{fig:teaser}
\vspace{-1.5em}
\end{figure}

To overcome the limitations of existing methods, we introduce a unified spatiotemporally decoupled framework with two specific modules as illustrated in~\cref{fig:teaser}, ensuring \textbf{high-fidelity generation} and \textbf{flexible controllability}.
For motion representation, we propose a lightweight spatial-temporal VAE that encodes each body joint rather than compressing the whole-body motion into a single latent space. The decoupled paradigm substantially enhances representation extraction performance and offers greater part-level controllability.
For the generative architecture, we incorporate spatial-temporal masking and attention mechanisms into an autoregressive diffusion generator, achieving both generative capability and controllable editability.

We conduct extensive experiments on HumanML3D~\cite{t2m} and KIT-ML~\cite{KIT} datasets to validate the superiority of our proposed spatiotemporally \textbf{De}coupled autoregressive \textbf{Mo}tion \textbf{Diff}usion model (DeMoDiff), for text-driven motion generation and motion editing. We also offer in-depth analysis and visualizations of our method, indicating that further advancements in autoregressive diffusion technology hold the potential to enhance its performance in the future.

In summary, our main contributions are as follows:

\begin{itemize}
    \item We novelly compress human motion into a continuous 2D latent space based on a spatial-temporal VAE, with improved reconstruction and generation metrics.

    \item A temporal-spatial masking strategy and diffusion mechanism adapted to continuous 2D latent of temporal-spatial representation, supporting fine-grained temporal-spatial control and motion generation.

    \item Our method demonstrates superior performance in several motion-related tasks, especially in joint-level editing.
\end{itemize}

%% file: sections/2_related_works_v1.tex
\section{Related Work}
\noindent \textbf{Text-driven Human Motion Generation.}
Text-driven human motion generation~\cite{t2m,temos,tmr} aims to synthesize realistic 3D human motions from natural languages, which poses significant challenges in fine-grained semantic mapping between texts and body movements, intricate body part control, and elaborate textual descriptions.

Vector Quantization (VQ)-based methods~\cite{TM2T,T2MGPT,motiongpt} encode motion data into discrete tokens for subsequent autoregressive motion synthesis. However, the sequential generation design inherently impedes the ability to model long-horizon temporal relationships.
Most recent works~\cite{momask,mogents} employ bidirectional masked generation models similar to BERT~\cite{bert}, which allows the model to predict all masked tokens simultaneously and enhances the context modeling capability.
Despite the impressive performance, VQ-based approaches suffer from inevitable information loss. In this paper, we introduce a lightweight spatial-temporal VAE to encode motion into latent vectors with better motion reconstruction performance.
Denoising diffusion models, which have achieved significant success in image synthesis, have now become powerful tools for generating human motion. Some attempts~\cite{MDM,MotionDiffuse,remodiffuse} operate on raw motion representation, while MLD~\cite{MLD} uses a Transformer-based VAE to compress the entire motion sequence to latent vectors.
Optimizing the denoising process in the latent space enhances training and sampling efficiency yet limits the model’s ability to capture temporal and spatial dynamics. 
Different from previous diffusion-based methods, we compress the raw motion into temporal and spatial latent space with a CNN-based encoder and decoder, and we autoregressively generate temporal and spatial motion latent during the inference stage with part-level control.

\noindent \textbf{Human Motion Representation.}
Human motion representation is critical for motion-related tasks. 
Previous works have extensively explored single human motion representations from the perspectives of both dimensional design and compact encoding.
From a dimensional perspective, HumanML3D~\cite{t2m} presents a 263-d representation that includes root angular velocity, root linear velocities, root height, local joint positions, velocities, rotations, and foot-ground contact.
MotionStreamer~\cite{motionstreamer} further introduces a 272-d representation to avoid data duplication and intricate post-processing of converting to SMPL parameters. 
MARDM~\cite{mardm} excludes the redundant dimensions of the 263-d format, resulting in a more essential and compact representation.
MotionPatches~\cite{motionpatches} divides and sorts skeleton joints based on body parts similar to image patches for ViT~\cite{dosovitskiy2020image} pre-training.
To achieve better joint encoding capability and joint-level spatial relationship, MoGenTS~\cite{mogents} splits the 263-d representation along the joint dimension.
From a compression perspective, many works~\cite{TM2T,T2MGPT,motiongpt,momask,mogents} represent motion as a discrete codebook. 
Some approaches~\cite{motionstreamer,mardm} mapping the motion sequence into a temporal latent sequence, while MLD~\cite{MLD} compresses the entire motion sequence into a single latent vector.
In this paper, we follow the dimensional setting of MoGenTS~\cite{mogents} for spatial-temporal decoupling and part-level editing, and subsequently encode motion into a 2D latent representation.

%% file: sections/3_method_v1.tex
\section{The DeMoDiff Model}
\begin{figure*}[t]
  \center
    \includegraphics[width=2\columnwidth, trim=0 0 0 0, clip]{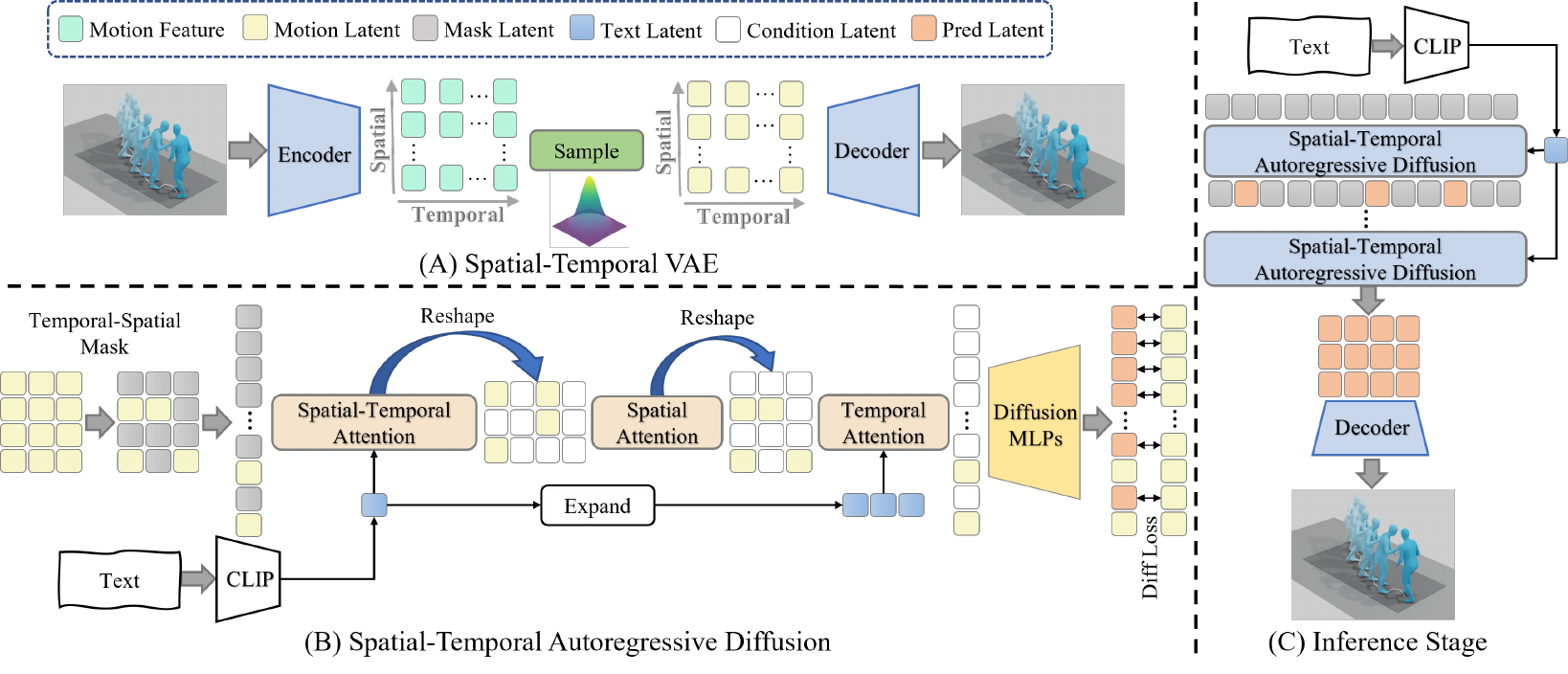}
    \vspace{-1.0em}
  \caption{\textbf{Overview of DeMoDiff.} (A) Spatial-Temporal VAE encodes motion into 2D latent and decodes it for reconstruction. (B) Spatial-Temporal Autoregressive Diffusion includes Spatial-Temporal Attention, Spatial Attention and Temporal Attention and is trained via mask modeling, using a diffusion head to predict motion latent. (C) During inference, Spatial-Temporal Autoregressive Diffusion iteratively predicts the latent representations, which are decoded into the motion sequence.}
  \label{fig2}
  \vspace{-1.75em}
\end{figure*}
The architecture overview of our proposed DeMoDiff model is demonstrated in~\cref{fig2}.
We provide a detailed explanation of spatial-temporal VAE for motion representation and spatial-temporal autoregressive diffusion architecture for continuous human motion generation.

\subsection{Spatial-Temporal VAE.}

The architecture of our proposed spatial-temporal VAE is shown in~\cref{fig2}~(A).
Following the previous method~\cite{mogents}, we organize the motion sequence into a 2D structure (in both temporal and joint dimensions) and process it as a 2D map, serving as the input to the VAE.
For a motion sequence $\mathbf{X}_{2d}=\{x^{i,k}_{2d}\}_{i=1,k=1}^{T,J}$ with length $T$ and joint number $J$, we employ a 2D convolutional network to encode the 2D motion features ${x}_{2d}$ into parameters of latent probability distributions, which are decoded to motion sequence. 
The network is optimized by a loss considering both reconstruction quality and latent space regularization, as:
\begin{equation}
    \mathcal{L}_{2d} = \mathcal{L}^{{2d}}_{nll} + \lambda^{2d}_{k} \cdot \mathcal{L}^{2d}_{KL} + \lambda_{v} \cdot \mathcal{L}_{v},
\end{equation}
where $\mathcal{L}^{{2d}}_{nll}$ is negative log-likelihood (NLL) loss, $\lambda^{2d}_{k}$ and $\lambda_{v}$ are the hyper-parameters for the Kullback Leibler (KL) loss $\mathcal{L}^{2d}_{KL}$ and joint velocity loss $\mathcal{L}_{v}$. More details are shown in the supplementary material.

\subsection{Spatial-Temporal Autoregressive Diffusion.}
In this section, we present Spatial-Temporal Autoregressive Diffusion, a diffusion-based framework for motion generation.
Given a text input and a 2D latent vector, Spatial-Temporal Autoregressive Diffusion uses an AdaLN~\cite{adaln} transformer network to predict the condition at the masked locations. Then the condition is set to the diffusion head for generating the predicted latent. There are three attentions performed in this transformer: spatial-temporal attention, spatial attention, and temporal attention.

\textbf{Attention mechanism.}
For the spatial-temporal attention, CLIP~\cite{CLIP} is employed to extract the text feature, resulting in a text latent $\mathbb{O}_\text{text}$.
The text features are injected into the transformer via AdaLN~\cite{adaln}, while the 2D latent vectors are directly augmented with 2D positional encodings~\cite{mogents}.

After the position encoding $\mathbf{P}$, we flatten the 2D latent vector to a 1D structure.
We perform spatial-temporal attention on this sequence to preserve spatial-temporal relationships.

Spatial attention focuses solely on spatial dimensions by treating time frames as separate batches, which computes relationships between joints within each time batch.

Temporal attention is similar to spatial attention, focusing solely on temporal dimensions by treating joints as separate batches, which computes time-step relationships within each joint batch.

Since we utilize the bidirectional model, the three attention modules can be applied respectively; however, we choose the integration version in which the three modules are applied sequentially for better performance. 

\textbf{Masking and training strategy.}
Our masking strategy adopts the same mask ratio schedule following~\cite{MaskGIT, mogents}.
This ratio is computed as:
\begin{equation}
    \gamma(\tau) = \cos(\frac{\pi \tau}{2}).
\end{equation}
During training, $\tau$ is uniformly sampled from $(0, 1)$ ($\tau \sim \mathbf{U}(0, 1)$) to generate a mask ratio $\gamma(\tau)$. Temporal masking then randomly masks $\gamma(\tau) \times t \times j$ latent, and spatial masking randomly masks $\gamma(\tau) \times j$ latent for each individual frame, both following this ratio.

For enhanced disturbance in masked prediction, BERT's remasking mechanism~\cite{bert} is incorporated: A latent vector ${z}_{2d}$ selected for masking is replaced by $\mathbb{O}_\text{mask}$ (80\% chance), a random noise $\mathcal{N}$ (10\% chance), or kept as is (10\% chance).

During training, a random mask is applied on the latent sequence $\mathbf{Z}_{2d}$ which is then processed by the transformer. 
After the transformer processing, we obtain the intermediate latent $\mathbf{C}=\{c^{um}_{2d}\}$ from unmasked latent vectors and the text latent $\mathbb{O}_\text{text}$, which serve as the condition for
the diffusion head (a small MLP) to predict masked latent $\hat{\mathbf{Z}}_{2d}=\{z^{m}_{2d}\}$.
Following~\cite{mar, DDPM}, the loss function is defined as:
\setlength{\abovedisplayskip}{7pt}
\begin{equation}
    \label{eq:diffusion_loss}
    \mathcal{L} = \mathbb{E}_{\epsilon,t} [||\epsilon-\epsilon_{\theta}(z^{m}_{2d}|t,c^{um}_{2d})||^2],
\end{equation}
where $t$ denotes the timestep of the noise schedule.

During inference, all latent vectors are initially set as mask latents. Text is first encoded via CLIP to obtain a semantic embedding, which is then fed into an AdaLN~\cite{adaln} modulation mechanism, which dynamically generates modulation parameters (scaling and shifting factors) for the AdaLN~\cite{adaln} transformer, thereby injecting text-conditioned signals into the transformer's normalization layers to guide its processing. The transformer, under the guidance of these text-derived conditions, predicts a set of conditional latents. These latents are input to diffusion MLPs, which further predict the predicted latents. This newly generated latent set is then inserted back into the original sequence, and the aforementioned process is iteratively repeated until the complete latent sequence is predicted. Unlike previous works~\cite{mardm,motionstreamer}, we utilize DDIM~\cite{ddim} to reduce the denoising step in our inference stage.

\begin{table*}[t]
\small
\centering
\caption{\textbf{Evaluation on the HumanML3D dataset (upper half) and KIT-ML dataset (lower half).} We repeat the evaluation for 20 times and report the average with 95\% confidence interval. $\downarrow$ means the lower, the better, while $\uparrow$ represents the opposite. $\rightarrow$ indicates the closer to the real motion the better. We use \textbf{bold} to indicate the best result and \underline{underscore} for the second best.}
\label{tab:eval_humanml}
\begin{tabular}{l l c c c c c c }
\toprule
Datasets&Methods & FID $\downarrow$ & Top1 $\uparrow$ & Top2 $\uparrow$ & Top3 $\uparrow$  & MM-Dist $\downarrow$ & Diversity $\rightarrow$ \\ 
\midrule
\multirow{6.5}{*}{\makecell[c]{HumanML3D}}
&Ground Truth & $0.002^{\pm .000}$ & $0.511^{\pm .003}$ & $0.703^{\pm .003}$ & $0.797^{\pm .002}$  & $2.974^{\pm .008}$ & $9.503^{\pm .065}$ \\ 

\cmidrule{2-8}

&MDM~\cite{MDM} & $0.544^{\pm .044}$ & $0.320^{\pm .005}$ & $0.498^{\pm .004}$ & $0.611^{\pm .007}$  & $5.566^{\pm .027}$ & $\mathbf{9.559}^{\pm .086}$ \\ 
&MLD~\cite{MLD} & $0.473^{\pm .013}$ & $0.481^{\pm .003}$ & $0.673^{\pm .003}$ & $0.772^{\pm .002}$  & $3.196^{\pm .010}$ & $9.724^{\pm .082}$ \\ 
&MotionDiffuse~\cite{MotionDiffuse} & $0.630^{\pm .001}$ & $0.491^{\pm .001}$ & $0.681^{\pm .001}$ & $0.782^{\pm .001}$  & $3.113^{\pm .001}$ & $\underline{9.410}^{\pm .049}$ \\ 
&ReMoDiffuse~\cite{remodiffuse} & $\mathbf{0.103}^{\pm .004}$ & $\underline{0.510}^{\pm .005}$ & $\underline{0.698}^{\pm .006}$ & $\underline{0.795}^{\pm .004}$  & $\underline{2.974}^{\pm .016}$ & $9.018^{\pm .075}$ \\ 

&DeMoDiff (Ours) & $\underline{0.164}^{\pm .012}$& $\mathbf{0.517}^{\pm .003}$ & $\mathbf{0.712}^{\pm .004}$ & $\textbf{0.808}^{\pm .002}$ & $\mathbf{2.907}^{\pm .013}$  & $9.399^{\pm .096}$  \\ 
\midrule
\multirow{6.5}{*}{\makecell[c]{KIT-ML}}
&Ground Truth & $0.031^{\pm .004}$ & $0.424^{\pm .005}$ & $0.649^{\pm .006}$ & $0.779^{\pm .006}$  & $2.788^{\pm .012}$ & $11.08^{\pm .097}$ \\ 
\cmidrule{2-8}

&MDM~\cite{MDM} & $0.497^{\pm .021}$ & $0.164^{\pm .004}$ & $0.291^{\pm .004}$ & $0.396^{\pm .004}$  & $9.191^{\pm .022}$ & $10.85^{\pm .109}$ \\ 
&MLD~\cite{MLD} & $0.404^{\pm .027}$ & $0.390^{\pm .008}$ & $0.609^{\pm .008}$ & $0.734^{\pm .007}$  & $3.204^{\pm .027}$ & $10.80^{\pm .117}$ \\ 
&MotionDiffuse~\cite{MotionDiffuse} & $1.954^{\pm .062}$ & $0.417^{\pm .004}$ & $0.621^{\pm .004}$ & $0.739^{\pm .004}$  & $2.958^{\pm .005}$ & $\mathbf{11.10}^{\pm .143}$ \\ 
&ReMoDiffuse~\cite{remodiffuse} & $\mathbf{0.155}^{\pm .006}$ & $\underline{0.427}^{\pm .014}$ & $\underline{0.641}^{\pm .004}$ & $\underline{0.765}^{\pm .055}$  & $\mathbf{2.814}^{\pm .012}$ & $10.80^{\pm .105}$ \\ 

&DeMoDiff (Ours)  & $\underline{0.263}^{\pm .017}$ & $\mathbf{0.432}^{\pm .008}$ & $\mathbf{0.660}^{\pm .007}$ & $\mathbf{0.785}^{\pm .006}$  & $\underline{2.827}^{\pm .019}$ & $\underline{10.86}^{\pm .112}$  \\ 
\bottomrule
\end{tabular}
\vspace{-1.5em}
\end{table*}

%% file: sections/4_experiment_v1.tex
\begin{figure*}[t]

  \center
    \includegraphics[width=2\columnwidth, trim=0 0 0 0, clip]{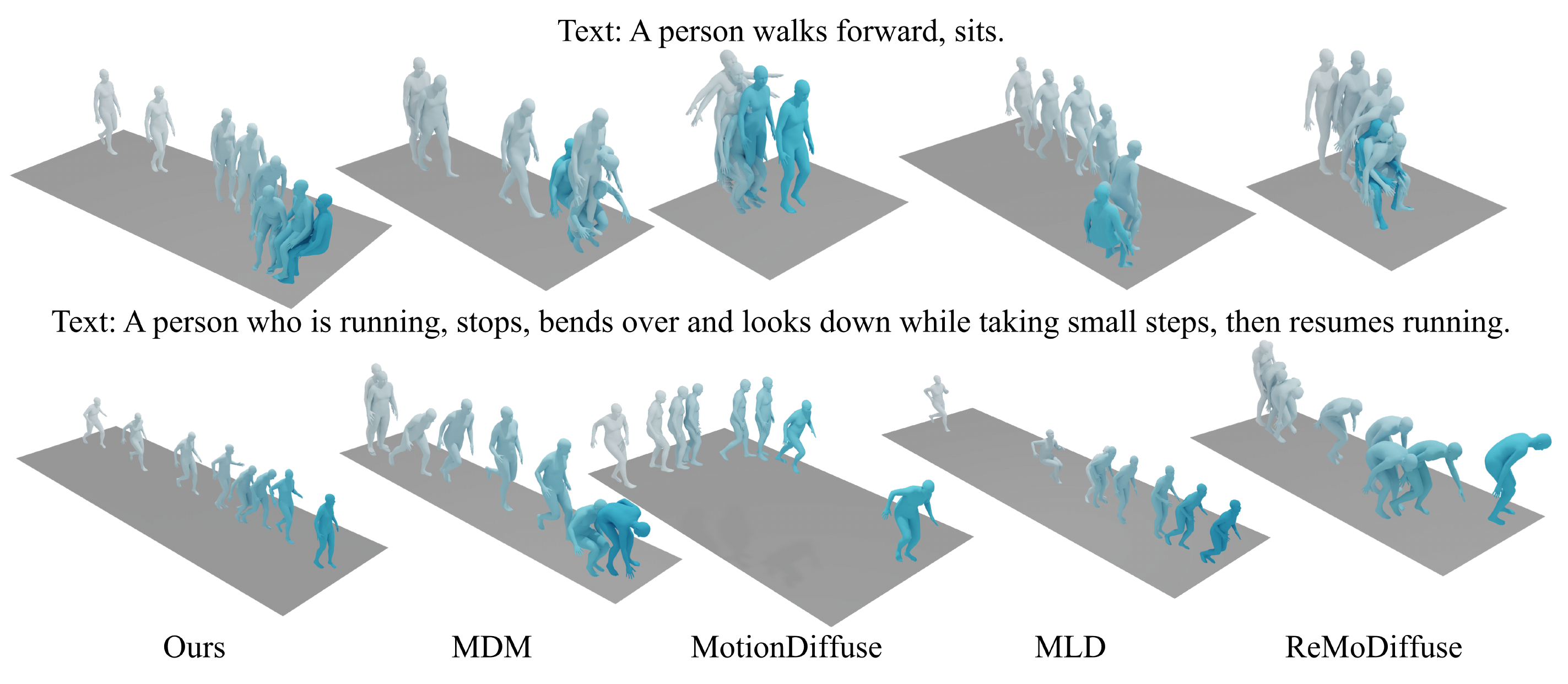}
    \vspace{-1.0em}
  \caption{\textbf{Visual comparison of generated results.} The color from light blue to dark blue indicates the motion sequential order.}
  \label{fig:vis_results}
  \vspace{-1.5em}
\end{figure*}

\section{Experiments}
\subsection{Datasets.}
We adopt two widely used text-to-motion datasets, \textit{i.e.}, HumanML3D~\cite{t2m} and KIT-ML~\cite{KIT}, to evaluate our proposed DeMoDiff model. 
HumanML3D includes 14,616 motion sequences and 44,970 corresponding text descriptions, while KIT-ML contains 3,911 motions and 6,278 texts. Following prior approaches, HumanML3D uses 23,384/1,460/4,383 samples for training, validation, and testing, respectively; KIT-ML employs 4,888/300/830 samples for the same splits. 
The human motions are extracted into features with 263 dimensions for HumanML3D and 251 dimensions for KIT-ML as in~\cite{t2m}, encompassing global elements like root velocity, root height, and foot contact, as well as local joint information (22 joints for HumanML3D and 21 joints for KIT-ML).

\subsection{Evaluation Metrics.}
\label{subsec:eval_metrics}
\begin{table}  
\centering
\caption{\textbf{Evaluation of motion compression methods on HumanML3D dataset.} MPJPE and PAMPJPE are measured in millimeters. ACCL indicates acceleration error.}
\label{tab:eval_vae}
\resizebox{\columnwidth}{!}{
\begin{tabular}{l c c c c c}
\toprule
Methods & \multicolumn{2}{c}{Generation} & \multicolumn{3}{c}{Reconstruction} \\
\cmidrule(lr){2-3} \cmidrule(lr){4-6}
 & FID $\downarrow$ & Diversity $\rightarrow$ & MPJPE $\downarrow$ & PA-MPJPE $\downarrow$ & ACCL $\downarrow$\\
 
\midrule
TM2T~\cite{TM2T} & 0.320 & 9.705 & 57.5 & 42.3 & 7.8 \\
T2M-GPT~\cite{T2MGPT} & 0.119 & 9.705 & 58.0 & 41.1 & 7.0 \\
MoMask~\cite{momask} & 0.019 & 9.662 & 29.5 & 20.3 & 5.3 \\
MoGenTS~\cite{mogents} & \textbf{0.005} & \textbf{9.547} & 16.4 & \underline{8.0} & \underline{3.0} \\
VPoser-t~\cite{vposer-t} & 1.430 & 8.336 & 75.6 & 48.6 & 9.3\\
ACTOR ~\cite{Action-Conditioned} & 0.341 & 9.569 & 65.3 & 41.0 & 7.0 \\
MLD~\cite{MLD} & 0.017 & \underline{9.554} & \underline{14.7} & 8.9 & 5.1 \\
DeMoDiff (Ours) & \textbf{0.005} & 9.413 & \textbf{9.4} & \textbf{2.7} & \textbf{2.4} \\
\bottomrule
\end{tabular}
}
\vspace{-2.0em}
\end{table}

We evaluate both Spatial-Temporal VAE and motion generation within our framework. The evaluation process strictly follows the standard procedures established by previous research~\cite{t2m,T2MGPT,momask}. 
Detailed explanation of these metrics are provided in the supplementary material.

We compare our model with previous diffusion-based text-to-motion works, including the continuous regression-based methods. From the results in~\cref{tab:eval_humanml}, our method achieves performance that is highly competitive with previous state-of-the-art approaches on both the HumanML3D and KIT-ML datasets, reflecting the effectiveness of our design. The full comparison results and analysis are provided in supplementary material.
DeMoDiff achieves better text-motion alignment, as evidenced by higher R-precision and lower MM-Dist.
The relatively lower FID of ReMoDiffuse is mainly attributed to its retrieval module, whose retrieved motion segments may deviate from the distribution of dataset-native motions, increasing distributional mismatch despite improvements in semantic alignment.
DeMoDiff consistently achieves comparable metrics, demonstrating its effectiveness in generating coherent and high-quality human motion.
We also demonstrate visualization comparison results in~\cref{fig:vis_results}. More visualization results of DeMoDiff are also provided in the supplementary material.

\begin{figure}
  \center
    \includegraphics[width=\columnwidth]{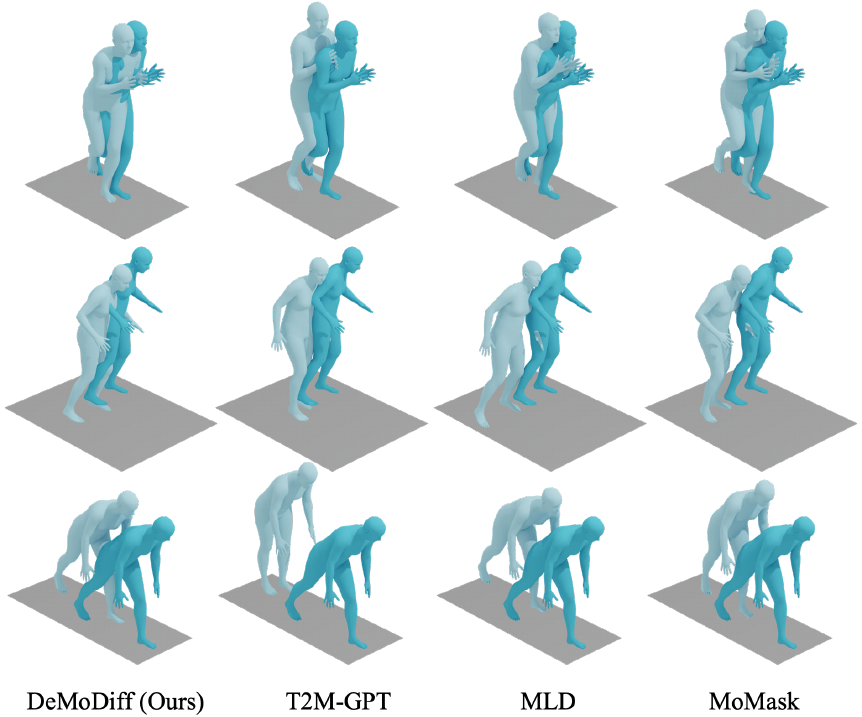}
  \caption{\textbf{Comparison of different motion compression methods.} The dark blue represents ground-truth, the light blue represents predicted results ( Note that the global position of SMPL is the global position of motion. We have not performed any additional alignment processing ).}
  \label{fig:vae_vis}
  \vspace{-1.5em}
\end{figure}

The core idea of our motion representation model is Spatial-Temporal VAE, which differs from previous quantization-based methods. Therefore, we compare with both previous continuous methods and quantization-based methods for a fair comparison. As reported in~\cref{tab:eval_vae}, the reconstruction accuracy of our method surpasses that of previous methods by a significant margin for the HumanML3D datasets. We also present visualization results with the SMPL body model in~\cref{fig:vae_vis}. It can be observed that our method outperforms other approaches in both motion similarity and global positional alignment. This reveals that the effect of our Spatial-Temporal VAE is much better than that of previous approaches.

\begin{table*} 
\small
\centering
\caption{\textbf{Ablation study of Spatial-Temporal VAE on the HumanML3D dataset.}}
\label{tab:ablation_klvae}
\resizebox{2\columnwidth}{!}{
\begin{tabular}{l c c c c c c c}
\toprule
Different VAE  & FID $\downarrow$ & Top1 $\uparrow$ & Top2 $\uparrow$ & Top3 $\uparrow$  & MM-Dist $\downarrow$ & Diversity $\rightarrow$ & MPJPE $\downarrow$\\ 
\midrule
1D Temporal (16 latent dim) & $0.0144^{\pm .0002}$ & $0.5071^{\pm .0024}$ & $0.6984^{\pm .0025}$ & $0.7936^{\pm .0020}$  & $3.0092^{\pm .0096}$ & $9.4839^{\pm .1060}$ & $28.0^{\pm .1}$\\ 
Ours (16 latent dim) & $0.0054^{\pm .0001}$ & $\mathbf{0.5142}^{\pm .0026}$ & $0.7015^{\pm .00025}$ & $\underline{0.7983}^{\pm .0024}$  & $\mathbf{2.9740}^{\pm .0085}$ & $9.5407^{\pm .0736}$ & $\underline{9.4}^{\pm .1}$ \\ 
Ours (32 latent dim) & $\underline{0.0048}^{\pm .0000}$ & $\underline{0.5126}^{\pm .0023}$ & $\mathbf{0.7042}^{\pm .0018}$ & $\mathbf{0.7988}^{\pm .0019}$  & $\underline{2.9757}^{\pm .0073}$ & $\mathbf{9.5081}^{\pm .0710}$ & $9.8^{\pm .1}$\\ 
Ours (64 latent dim) & $\mathbf{0.0041^{\pm .0001}}$ & $0.5111^{\pm .0020}$ & $\underline{0.7019}^{\pm .0018}$ & $0.7951^{\pm .0016}$  & $2.9806^{\pm .0067}$ & $\underline{9.4896}^{\pm .0716}$ & $\mathbf{7.7^{\pm .1}}$\\ 
\bottomrule
\end{tabular}
}
\vspace{-0.5em}
\end{table*}

\begin{table*} 
\small
\centering
\caption{\textbf{Ablation study of the attention mechanisms choices on the HumanML3D dataset.}}
\label{tab:eval_attention}
\begin{tabular}{l c c c c c c }
\toprule
Methods & FID $\downarrow$ & Top1 $\uparrow$ & Top2 $\uparrow$ & Top3 $\uparrow$  & MM-Dist $\downarrow$ & Diversity $\rightarrow$ \\ 
\midrule
1D Temporal VAE + temporal attention & $0.665^{\pm .017}$ & $0.452^{\pm .003}$ & $0.633^{\pm .004}$ & $0.735^{\pm .003}$  & $3.361^{\pm .011}$ & $9.114^{\pm .070}$ \\   
\midrule
w/o all attention & $37.769^{\pm .122}$ & $0.086^{\pm .003}$ & $0.156^{\pm .004}$ & $0.212^{\pm .003}$  & $7.511^{\pm .011}$ & $3.374^{\pm .083}$ \\ 

w/o temporal\&spatial attention & $1.315^{\pm .034}$ & $0.443^{\pm .007}$ & $0.639^{\pm .007}$ & $0.748^{\pm .006}$  & $3.269^{\pm .013}$ & $8.905^{\pm .147}$ \\ 

w/o spatial attention & $0.783^{\pm .015}$ & $0.463^{\pm .003}$ & $\underline{0.661}^{\pm .004}$ & $\underline{0.765}^{\pm .007}$  & $3.212^{\pm .022}$ & $\underline{9.232}^{\pm .127}$ \\ 

w/o temporal attention & $\underline{0.752}^{\pm .026}$ & $\underline{0.466}^{\pm .005}$ & $0.657^{\pm .004}$ & $0.761^{\pm .004}$  & $\underline{3.168}^{\pm .018}$ & $9.193^{\pm .095}$ \\ 

DeMoDiff (Full Model) & $\mathbf{0.164}^{\pm .012}$ & $\mathbf{0.517}^{\pm .003}$ & $\mathbf{0.712}^{\pm .004}$ & $\mathbf{0.808}^{\pm .002}$  & $\mathbf{2.907}^{\pm .013}$ & $\mathbf{9.399}^{\pm .096}$ \\ 
\bottomrule
\end{tabular}
\vspace{-1.5em}
\end{table*}
\begin{figure*}
  \center
    \includegraphics[width=2\columnwidth, trim=0 0 0 0, clip]{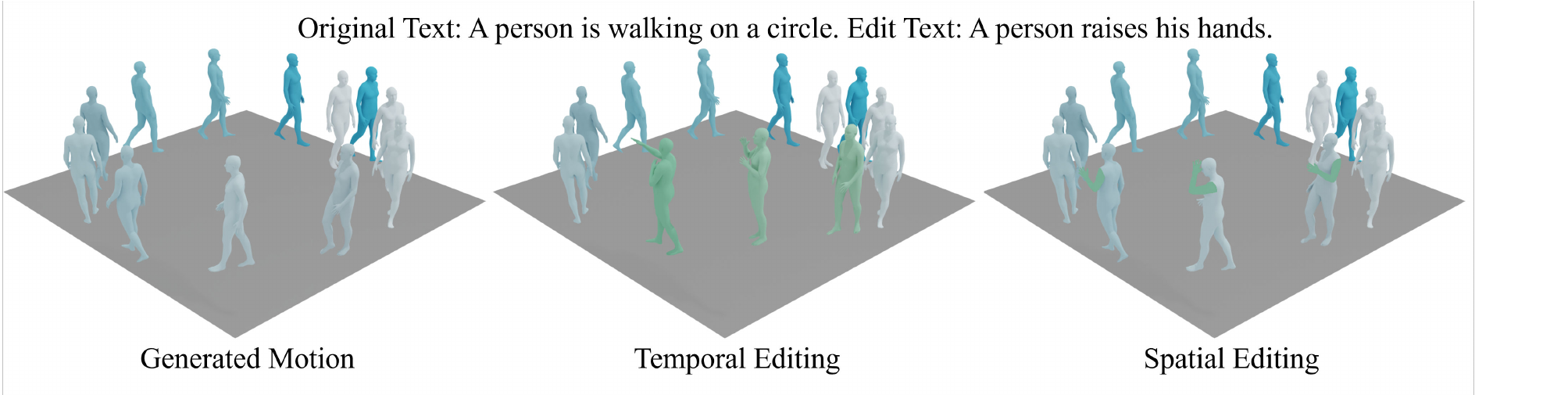}
    \vspace{-1.0em}
  \caption{\textbf{Visualization of editing results.} The color from light blue to dark blue indicates the motion sequence order. The green color indicates the edited regions.}
  \label{fig:editing_results}
  \vspace{-1.5em}
\end{figure*}
\subsection{Ablation Study.}
\label{subsection:Ablation Study}
To verify the efficacy of the proposed framework, we conducted ablation experiments on Spatial-Temporal VAE and Spatial-Temporal Autoregressive Diffusion. 
For the VAE, we first establish a baseline model incorporating a 1D Temporal motion VAE. 
Then, we change the VAE to our Spatial-Temporal VAE, examining its impact on both motion representation and motion generation capabilities. We also evaluate the influence of different VAE latent sizes. Regarding Spatial-Temporal Autoregressive Diffusion, we assessed the impact of various attention choices on human motion generation. 

\textbf{Spatial-Temporal VAE.}
As shown in~\cref{tab:ablation_klvae}, we conducted a comparative analysis between our Spatial-Temporal VAE and the 1D Temporal VAE, alongside an investigation into the effects of varying VAE latent dimension sizes. 
Our experiments reveal that a larger latent dimension in the VAE leads to improvements in reconstruction metrics. This can be attributed to the fact that a higher-dimensional latent space provides a more abundant representational capacity for data distribution, enabling it to capture the detailed features of input data more precisely.

\textbf{Spatial-Temporal Autoregressive Diffusion.}
We analyze the impact of different design choices of the attention mechanisms, as shown in~\cref{tab:eval_attention}. These attention mechanisms for Spatial-Temporal Autoregressive Diffusion were evaluated both in their individual applications and in sequential stacked configurations. We first removed all attention modules from the model, and then incrementally stacked the temporal-spatial attention, temporal attention, and spatial attention modules in a step-by-step manner until the complete model architecture was reconstructed. We also evaluate the metrics of 1D temporal VAE and temporal attention. Experimental results demonstrate that the model achieves a significant performance improvement when all three attention modules are integrated, which validates the effectiveness of the proposed model design.

\subsection{Temporal and Spatial Editing.}
Since we utilize bidirectional attention in our framework, our method supports both temporal editing and spatial editing without requiring any editing-specific fine-tuning, which differs from previous diffusion-based works~\cite{MLD,MDM,MotionDiffuse,remodiffuse,motionstreamer,mardm}. For temporal motion editing, we treat all latents corresponding to the temporal dimensions that need to be edited as mask latents and then generate motions following our standard generation procedure, which conditions on the unmasked latents and the editing textual instructions. For spatial editing, similar to the temporal editing step, we only treat the latents of the spatial dimension to be edited as mask latents.
We demonstrate the visualization results in~\cref{fig:editing_results}

%% file: sections/5_conclusion_v1.tex
\section{Conclusion}

We present DeMoDiff as a novel framework for text-to-motion generation that
integrates a spatiotemporally decoupled autoregressive diffusion model to directly predict temporal and spatial motion latents. By introducing the spatial-temporal VAE and the spatial-temporal autoregressive diffusion model,
DeMoDiff enables sampling in both temporal and spatial latent spaces, with the ability for temporal and spatial editing. Our method demonstrates its competitiveness in motion generation while providing greater flexibility.